\documentclass{article}
\usepackage{spconf,amsmath,epsfig}

\usepackage{subfig}
\usepackage{amssymb}
\usepackage{booktabs}
\usepackage[switch]{lineno}
\usepackage{multirow}
\usepackage[pagebackref=false,breaklinks=true,colorlinks]{hyperref}
\let\OLDthebibliography\thebibliography
\renewcommand\thebibliography[1]{
  \OLDthebibliography{#1}
  \setlength{\parskip}{0pt}
  \setlength{\itemsep}{0pt plus 0.3ex}
}

\begin{document}\sloppy

\def\x{{\mathbf x}}
\def\L{{\cal L}}

\title{BargainNet: Background-Guided Domain Translation for Image Harmonization}
%
\name{Wenyan Cong, Li Niu\textsuperscript{*}\thanks{\textsuperscript{*}Corresponding author.}, Jianfu Zhang, Jing Liang, Liqing Zhang}
\address {MoE Key Lab of Artificial Intelligence, Department of Computer Science and Engineering \\
Shanghai Jiao Tong University, Shanghai, China \\ 
\{plcwyam17320, ustcnewly, c.sis, leungjing\}@sjtu.edu.cn, zhang-lq@cs.sjtu.edu.cn.}

\maketitle

\begin{abstract}
Given a composite image with inharmonious foreground and background, image harmonization aims to adjust the foreground to make it compatible with the background. Previous image harmonization methods mainly focus on learning the mapping from composite image to real image, while ignoring the crucial guidance role that background plays. In this work,  we formulate image harmonization task as background-guided domain translation. Specifically, we use a domain code extractor to capture the background domain information to guide the foreground harmonization,  which is regulated by well-tailored triplet losses. Extensive experiments on the benchmark dataset demonstrate the effectiveness of our proposed method. Code is available at \href{https://github.com/bcmi/BargainNet}{https://github.com/bcmi/BargainNet}. 
\end{abstract}
\vspace{-0.2cm}
\section{Introduction}

\begin{figure*}[tp!]
\setlength{\abovecaptionskip}{-0.2cm}
\setlength{\belowcaptionskip}{-0.4cm}
\begin{center}
\includegraphics[width=0.85\linewidth]{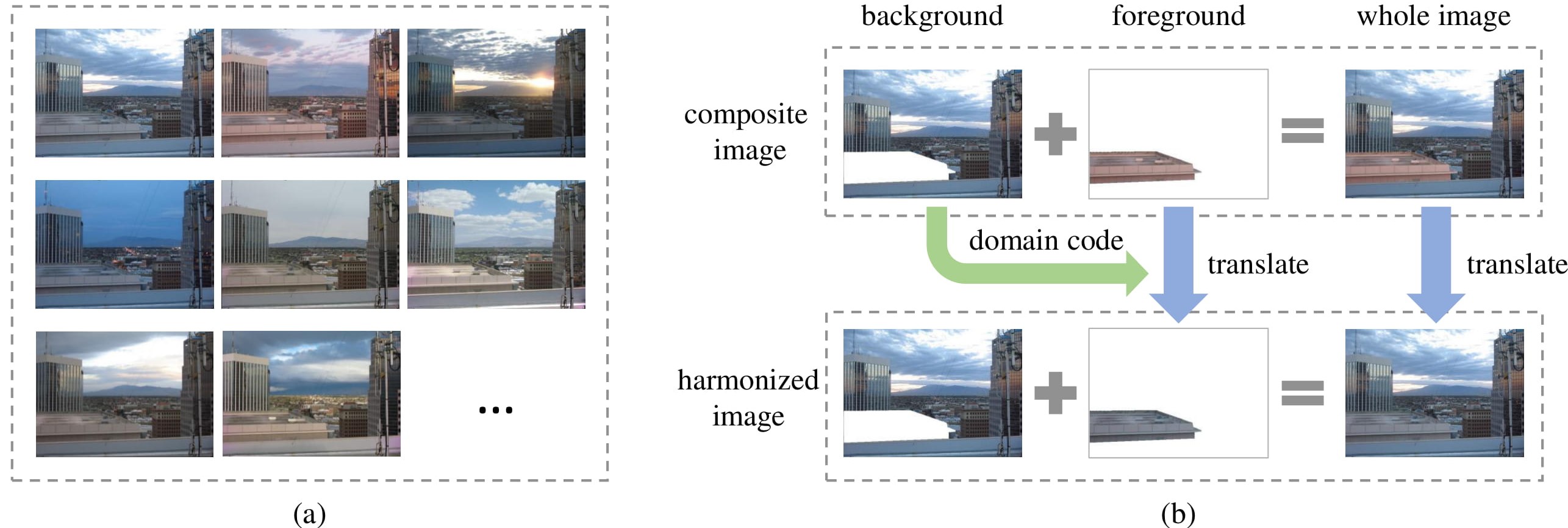}
\end{center}
  \caption[]{(a) Illustration of different domains corresponding to different capture conditions. (b) Our BargainNet utilizes background domain code to guide the foreground domain translation, resulting in consistent foreground and background.}
  \label{fig:intro}

\end{figure*}

Image composition synthesizes the composite by combining the foreground from one image with the background from another image. One issue of image composition is the appearance differences between foreground and background caused by distinct capture conditions (\emph{e.g.}, weather, season, time of day). Therefore, making the generated composite realistic could be a challenging task. Image harmonization~\cite{tsai2017deep,xiaodong2019improving,DoveNet2020}, which aims to adjust the foreground to make it compatible with the background, is essential to address this problem. 
Traditional harmonization methods~\cite{lalonde2007using,xue2012understanding,multi-scale} improve the quality of synthesized composite mainly by transferring hand-crafted appearance statistics between foreground and background regions, but they could not handle the large appearance gap between foreground and background regions. Recently, more deep learning based harmonization approaches have also been proposed. In \cite{tsai2017deep}, they presented the first end-to-end network for image harmonization. In \cite{xiaodong2019improving}, the spatial-separated attention blocks were proposed to learn the foreground and background features separately. Later in \cite{DoveNet2020}, they proposed an adversarial network with a domain verification discriminator to pull close the domains of foreground and background regions. Nonetheless, previous deep learning based methods neglected the crucial guidance role that background plays in the harmonization task. Therefore, they did not realize the shortcut to addressing image harmonization by posing it as background-guided domain translation. 

According to DoveNet~\cite{DoveNet2020}, we can treat different capture conditions as different domains. As illustrated in Fig.~\ref{fig:intro}(a), there could be innumerable possible domains for natural images. Even for the same scene, when the season, weather, time of the day, or photo equipment settings vary, the domain changes. 
For a real image, its foreground and background are captured in the same condition and thus belong to the same domain. But for a composite image, its foreground and background may belong to two different domains. In this case, image harmonization could be regarded as transferring the foreground domain to the background domain, making it a special case of domain translation. Domain translation has been extensively explored in \cite{pixelGAN,cycleGAN,starGan,lee2020drit++,choi2020stargan,f2gan,matchinggan}, and most domain translation methods require explicitly predefined domain labels, which are unavailable in our task. More recently, methods without domain labels have also been proposed as exemplar-guided domain translation~\cite{anokhin2020high,wang2019example}, in which an exemplar image provides the domain guidance. 

In this paper, we take a further step beyond exemplar-guided domain translation and detail the problem to local region guidance, \emph{i.e.}, background-guided domain translation. As demonstrated in Fig.~\ref{fig:intro}(b), the background and foreground of a composite image belong to different domains. With the guidance of extracted background domain code, which encodes the domain information of background, the composite foreground could be translated to the same domain as background, leading to a harmonious output.

As we propose to address image harmonization problem from a new perspective, one of our main contributions is the proposed \textbf{Ba}ckg\textbf{r}ound-\textbf{g}uided dom\textbf{ain} translation \textbf{Net}work, which is called BargainNet for short. Since partial convolution \cite{Liu2018} only concentrates on the feature aggregation of a partial region, we leverage partial convolution in our domain code extractor to extract the background domain information, which can avoid the information leakage between foreground and background. The obtained background domain code defines the target domain and guides the foreground domain translation. There are various ways of utilizing the target domain code to guide domain translation. For simplicity, we spatially replicate the background domain code to the same size as input image and concatenate them along the channel dimension. The concatenated input, together with the foreground mask, is fed into an attention-enhanced U-net generator~\cite{DoveNet2020} to produce the harmonized result. At the same time, we propose two well-tailored triplet losses to ensure that the domain code extractor can indeed extract domain information instead of domain-irrelevant information (\emph{e.g.}, semantic layout). The proposed triplet losses pull close the domain codes of background, real foreground, and the harmonized foreground, while pushing the domain code of composite foreground apart from them. To verify the effectiveness of our proposed BargainNet, we conduct comprehensive experiments on the image harmonization dataset iHarmony4~\cite{DoveNet2020}.



The contributions of our method are four-fold. 1) To the best of our knowledge, we are the first to formulate the image harmonization task as background-guided domain translation, which provides a new perspective for image harmonization; 2) We propose a novel image harmonization network, \emph{i.e.}, BargainNet, equipped with domain code extractor and well-tailored triplet losses; 3) Our method can extract meaningful domain code, which has other potential usages like inharmony level prediction; 4) Our method achieves the competitive performance on the benchmark dataset.

\vspace{-0.2cm}
\section{Related Work}


\textbf{Image Harmonization: }Image harmonization aims to make the composite foreground compatible with the background. To adjust the foreground appearance, traditional methods mainly leveraged low-level appearance statistics~\cite{reinhard2001color,colorharmonization,poisson,multi-scale}. Later in \cite{lalonde2007using,xue2012understanding,zhu2015learning}, image realism was gradually explored to make the composite image more realistic.

Recently, harmonization methods that synthesize paintings from photo-realistic images have been explored in \cite{luan2018deep,shaham2019singan}. However, they are more like style transfer and different from the photo-realistic harmonization in our task. More related to our work, in \cite{tsai2017deep,xiaodong2019improving,DoveNet2020}, they directly learn a mapping from composite images to real images, with the assistance of auxiliary semantic parsing branch~\cite{tsai2017deep}, inserted attention models~\cite{xiaodong2019improving}, or domain verification discriminator~\cite{DoveNet2020}. Different from these existing methods, our proposed method provides a new perspective by treating image harmonization as a background-guided domain translation.

\textbf{Domain Translation: }The task of domain translation aims to learn the mapping from a source domain to a target domain (\emph{e.g.}, from day to night). Recent works could be divided into two main streams: methods that require domain labels ~\cite{pixelGAN,cycleGAN,huang2018munit,lee2020drit++,choi2020stargan,f2gan,matchinggan} and methods without any predefined domain labels~\cite{anokhin2020high,wang2019example,ma2018exemplar}. 
In image harmonization, domains correspond to different capture conditions. Therefore, domain labels are hard to define and hard to solicit from users. So our work is more related to the latter, which is also known as example-guided domain translation. Given an exemplar image as guidance, the input image is translated into the same domain as the given exemplar image. In this paper, we take a further step and pose image harmonization as background-guided domain translation, which utilizes background region instead of an exemplar image as guidance. 
\vspace{-0.2cm}
\section{Our Method}

\begin{figure}[tp!]
\setlength{\abovecaptionskip}{-0.1cm}
\begin{center}
\includegraphics[width=\linewidth]{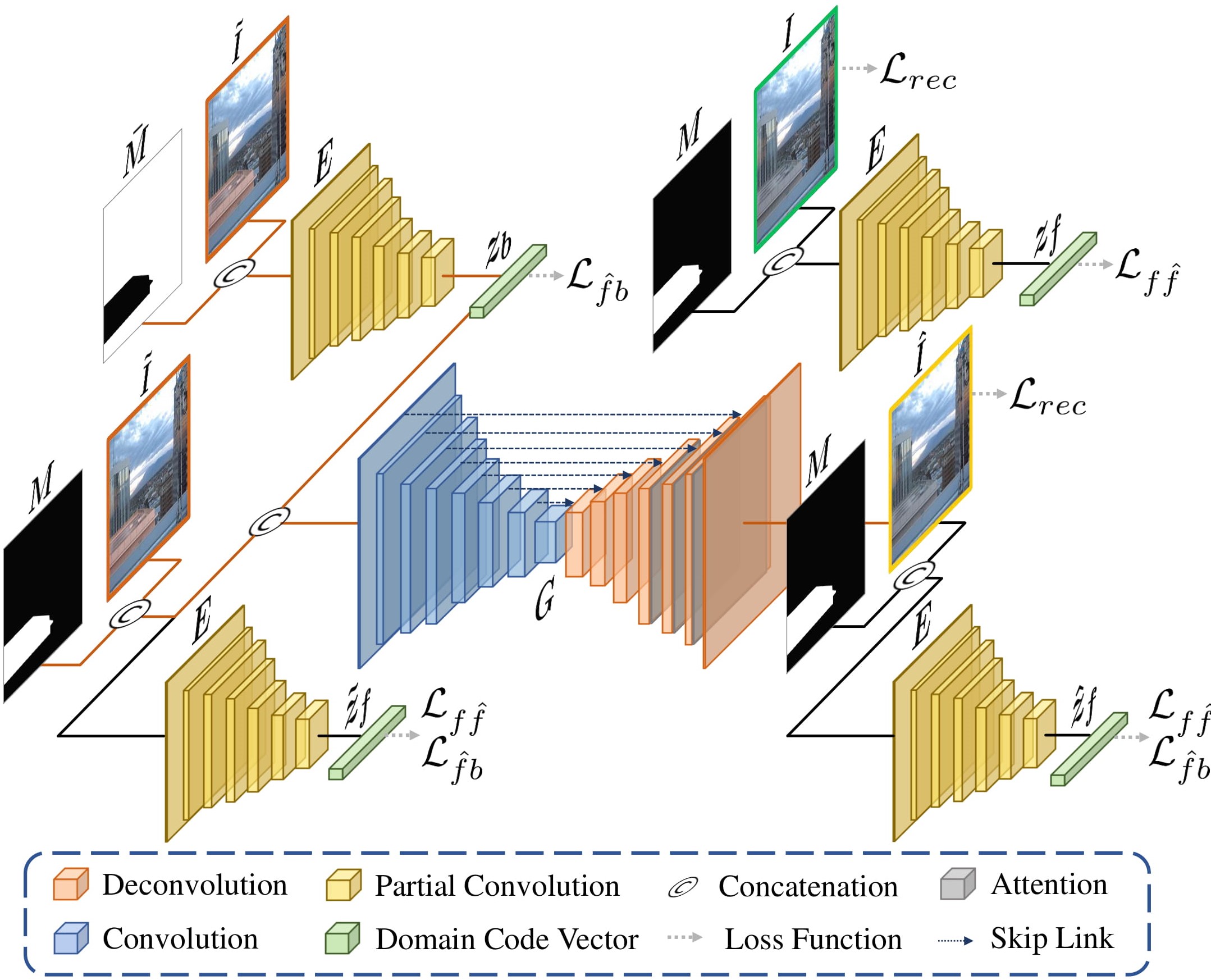}
\end{center}
   \caption{The network architecture of our BargainNet, which consists of attention enhanced U-Net generator $G$ and domain code extractor $E$. We employ two types of triplet losses based on four types of domain codes (see Section~\ref{sec:bargainnet}). The test phase is highlighted with red flow lines for clarity.}
\label{fig:flowchart}
\vspace{-0.4cm}
\end{figure}

In image harmonization task, we utilize training pairs of composite image $\tilde I\in \mathbb{R}^{H\times W\times 3}$ and real image $I \in \mathbb{R}^{H \times W \times 3}$, in which $H$ (\emph{resp.}, $W$) is image height (\emph{resp.}, width). The background of $I$ (real background) is the same as the background of $\tilde I$ (composite background). So in the remainder of this paper, we only mention background without distinguishing between real background and composite background. The foreground of $I$ (real foreground) is the harmonization target of the foreground of $\tilde I$ (composite foreground). The binary mask $M\in \mathbb{R}^{H \times W \times 1}$ indicates the foreground region to be harmonized, and therefore the background mask is $\bar{M}=1-M$. 

Given a composite image $\tilde I$, the goal of image harmonization task is to use a generator to reconstruct $I$ with a harmonized output $\hat I$, in which the foreground of $\hat I$ (harmonized foreground) should be close to the real foreground.
Next, we first introduce our domain code extractor in Section~\ref{sec:code_extractor}, and then introduce our whole network BargainNet in Section~\ref{sec:bargainnet}.

\vspace{-0.3cm}
\subsection{Domain Code Extractor}\label{sec:code_extractor}

To extract the domain code for a region with an irregular shape, our domain code extractor $E$ is composed of contiguously stacked partial convolutional layers \cite{Liu2018}, which are designed for special image generation with irregular masks. 
The output of the domain code extractor only depends on the aggregated features within the masked region, which prevents information leakage from the unmasked region.
For the technical details of partial convolution, please refer to \cite{Liu2018}.

In our task, we use domain code extractor to extract the domain codes of the foreground/background regions of composite image $\tilde I$, real image $I$, and output image $\hat I$. 
For example, given a composite image $\tilde I$ and its background mask $\bar M$, $E$ could extract the background domain code of $\tilde I$. To enforce the domain code to contain domain information instead of other domain-irrelevant information (\emph{e.g.}, semantic layout), we use background domain code to guide the foreground domain translation and design well-tailored triplet losses to regulate the domain code, which will be introduced next.

\vspace{-0.2cm}
\subsection{Background-guided Domain Translation Network} \label{sec:bargainnet}
Our proposed \textbf{Ba}ckg\textbf{r}ound-\textbf{g}uided dom\textbf{ain} translation \textbf{Net}work (BargainNet) has two modules: domain code extractor $E$ and  generator $G$. We adopt attention-enhanced U-net proposed in \cite{DoveNet2020} as $G$ and omit the details here.


As demonstrated in Fig. \ref{fig:flowchart}, given a composite image $\tilde I$ and its background mask $\bar M$, the domain code extractor takes $\tilde I$ and $\bar M$ as input and outputs the background domain code $z_b$. The extracted background domain code is used as the target domain code for foreground domain translation, which means that the foreground will be translated to the background domain with its domain-irrelevant information (\emph{e.g.}, semantic layout) well-preserved. Besides, the background should remain unchanged if we translate it to the background domain. So for ease of implementation, we simply translate both foreground and background to the background domain. 
Inspired by domain translation methods \cite{zhu2017toward,starGan}, we spatially replicate the $L$-dimensional domain code $z_b$ to an $H \times W \times L$ domain code map $Z_b$ and concatenate it with the $H \times W \times 3$ composite image. Besides, based on our experimental observation (see Section~\ref{sec:ablate} and Supplementary), it is still necessary to use foreground mask to indicate the foreground region to be harmonized as in ~\cite{tsai2017deep,xiaodong2019improving,DoveNet2020}, probably because the foreground mask emphasizes foreground translation and enables the foreground to borrow information from the background.
Thus, we further concatenate the input with the  $H \times W \times 1$ foreground mask $M$, leading to the final
$H \times W \times (L+4)$ input. After passing the input through the generator $G$, we enforce the harmonized output $\hat I=G(\tilde I, M, Z_b)$ to be close to the ground-truth real image $I$ by using the reconstruction loss $\mathcal{L}_{rec} = \|\hat I-I\|_1$.

We assume that $z_b$ only contains the domain information of background. Because if $z_b$ contains the domain-irrelevant information (\emph{e.g.}, semantic layout) of background, it may corrupt the semantic layout of foreground, which violates the reconstruction loss. To further reinforce our assumption on domain code, we use triplet losses to pull close the domain codes which are expected to be similar and push apart those which are expected to be divergent. Analogous to extracting background domain code $z_b$, we also use $E$ to extract the domain codes of real foreground, composite foreground, and harmonized foreground, denoted as $z_f$, $\tilde{z}_f$, and $\hat{z}_f$ respectively. For ease of description, we define an image triplet as a composite image, its ground-truth real image, and its harmonized output. Given an image triplet, we can obtain $\tilde z_f$, $z_b$, $z_f$ and $\hat z_f$. 

First, after harmonization, the foreground is translated from composite foreground domain to background domain. Hence, the domain code of harmonized foreground ($\hat z_f$) should be close to that of background ($z_b$), but far away from that of composite foreground ($\tilde z_f$).
In other words, we aim to pull close $\hat z_f$ and $z_b$  while pushing apart $\hat z_f$ and $\tilde z_f$, which can be achieved by the following triplet loss:
\begin{eqnarray} \label{eqn:triplet_fb}
\mathcal{L}_{\hat fb}\!\!\!\!\!\!\!\!&&= \mathcal{L}(\hat z_f,z_b,\tilde z_f)\\
&&=\max (d(\hat z_f, z_b)-d(\hat z_f, \tilde z_f)+m, 0),\nonumber
\end{eqnarray}
in which $d(\cdot,\cdot)$ is Euclidean distance and $m$ is a margin. 

\begin{table*}[tb]
\centering
\begin{tabular}{|l|c|c|c|c|c|c|c|c|c|c|}
\hline
\multicolumn{1}{|c|}{Sub-dataset} & \multicolumn{2}{c|}{HCOCO} & \multicolumn{2}{c|}{HAdobe5k} & \multicolumn{2}{c|}{HFlickr} & \multicolumn{2}{c|}{Hday2night} & \multicolumn{2}{c|}{All}\\ \hline
\multicolumn{1}{|c|}{Evaluation metric}  & MSE$\downarrow$  & PSNR$\uparrow$   & MSE$\downarrow$  & PSNR$\uparrow$ & MSE$\downarrow$  & PSNR$\uparrow$ & MSE$\downarrow$  & PSNR$\uparrow$  & MSE$\downarrow$  & PSNR$\uparrow$  \\ \hline
\multicolumn{1}{|c|}{Input composite}  & 69.37  & 33.94   & 345.54  & 28.16 & 264.35 & 28.32 & 109.65  & 34.01 & 172.47 & 31.63  \\ \hline
\multicolumn{1}{|c|}{Lalonde and Efros\cite{lalonde2007using}}  & 110.10  & 31.14   & 158.90  & 29.66 & 329.87  & 26.43 & 199.93 & 29.80  & 150.53 & 30.16 \\ \hline
\multicolumn{1}{|c|}{Xue \emph{et al.}\cite{xue2012understanding}}  & 77.04  & 33.32   & 274.15  & 28.79 & 249.54  & 28.32 & 190.51 & 31.24  & 155.87 & 31.40 \\ \hline
\multicolumn{1}{|c|}{Zhu \emph{et al.}\cite{zhu2015learning}}  & 79.82  & 33.04   & 414.31  & 27.26 & 315.42  & 27.52 & 136.71 & 32.32  & 204.77 & 30.72 \\ \hline
\multicolumn{1}{|c|}{DIH~\cite{tsai2017deep}}  & 51.85  & 34.69   & 92.65  & 32.28 & 163.38  & 29.55 & 82.34 & 34.62 & 76.77 & 33.41  \\ \hline
\multicolumn{1}{|c|}{DoveNet~\cite{DoveNet2020}} & 36.72 & 35.83 & 52.32  & 34.34 & 133.14 & 30.21 & 54.05 & 35.18 & 52.36 & 34.75  \\ \hline
\multicolumn{1}{|c|}{S$^2$AM~\cite{xiaodong2019improving}}  & 33.07 & 36.09  & 48.22 & \bf35.34 & 124.53  & 31.00 & \bf48.78 & 35.60 & 48.00 & 35.29  \\ \hline
\multicolumn{1}{|c|}{Ours}  & \bf24.84 & \bf37.03 & \bf39.94 & \bf35.34 & \bf97.32  & \bf31.34 & 50.98 & \bf35.67 & \bf37.82& \bf35.88\\ \hline
\end{tabular}
\caption{Quantitative comparison between our proposed BargainNet and other baseline methods. The best results are denoted in boldface.}
\label{tab:baselines}
\vspace{-0.3cm}
\end{table*}

Next, we consider the relationship among three foregrounds in an image triplet. 
The domain code of real foreground ($z_f$) should be close to that of harmonized foreground ($\hat z_f$), but far away from that of composite foreground ($\tilde z_f$). This goal can be achieved by the following triplet loss:
\begin{eqnarray} \label{eqn:triplet_ff}
\mathcal{L}_{f\hat f}\!\!\!\!\!\!\!\!&&=\mathcal{L}(z_f,\hat z_f,\tilde z_f)\\
&&=\max (d(z_f, \hat z_f)-d(z_f, \tilde z_f)+m, 0).\nonumber
\end{eqnarray}

In fact, there could be many reasonable combinations of triplet losses to regulate the domain code. However, based on our experimental observation, a combination of (\ref{eqn:triplet_fb}) and (\ref{eqn:triplet_ff}) has already met all our expectations (see Section~\ref{sec:domain_code}).
So far, the overall loss function for our method is
\begin{eqnarray} \label{eqn:total_loss}
    \mathcal{L}\!\!\!\!\!\!\!\!&&=\mathcal{L}_{rec}+\lambda \mathcal{L}_{tri} =\mathcal{L}_{rec}+\lambda(\mathcal{L}_{f\hat f}+\mathcal{L}_{\hat f b}),
\end{eqnarray}
where $\lambda$ is a trade-off parameter.


\vspace{-0.2cm}
\section{Experiments}

\begin{table*}
\centering
\begin{tabular}{|c|c|c|c|c|c|c|c|c|}
\hline
 & & $d_{b,f}<d_{b,\tilde f}$ & $d_{b,\hat f}<d_{b,\tilde f}$ & $d_{f,\hat f}<d_{f,\tilde f}$ & $d_{\hat f, f}<d_{\hat f,\tilde f}$ & $d_{\hat f, b}<d_{\hat f,\tilde f}$ & $d_{f, b}<d_{f,\tilde f}$ & All\\
\hline
\multirow{2}*{DoveNet\cite{DoveNet2020}} &Train & 47.08\% & 49.24\% & 72.22\% & 71.47\% & 12.01\% & 11.75\% & 5.93\% \\
\cline{2-9}
~ & Test  & 51.34\% &51.58\% &62.34\% &54.65\% &13.68\% &15.64\% & 5.09\% \\
\hline
\multirow{2}*{Ours} &Train  & 88.63\% & 97.87\% & 93.65\% & 91.92\% & 96.38\% & 87.98\% & 80.70\% \\
\cline{2-9}
~ & Test  & 90.28\% &97.39\% &91.87\% &89.28\% &96.26\% &89.09\% & 81.36\% \\
\hline
\end{tabular}
\caption{The ratio of training/testing image triplets which satisfy the specified requirements of DoveNet and our method. Note that $d_{x,y}$ is short for $d(z_x, z_y)$. For example, $d_{b,f}$ denotes the Euclidean distance between the background domain code $z_b$ and the domain code of real foreground $z_f$.}
\label{tab:domain_code}
\vspace{-0.4cm}
\end{table*}

\vspace{-0.1cm}
\subsection{Dataset and Implementation Details}
We evaluate our method and baselines on the benchmark dataset iHarmony4~\cite{DoveNet2020}, which contains 73146 pairs of synthesized composite images and the ground-truth real images (65742 pairs for training and 7404 pairs for testing). iHarmony4 consists of four sub-datasets: HCOCO, HAdobe5k, HFlickr, and Hday2night. The details of four sub-datasets can be found in the Supplementary.

The extracted domain code is a 16-dimension vector. We set the margin $m$ in Eqn. (\ref{eqn:triplet_fb})(\ref{eqn:triplet_ff}) as 1 and the trade-off parameter $\lambda$ in Eqn. (\ref{eqn:total_loss}) as 0.01.
In our experiments, the input images are resized to $256\times 256$ during both training and testing phases. 
Following \cite{tsai2017deep,DoveNet2020}, we use Mean-Squared Errors (MSE) and Peak Signal-to-Noise Ratio (PSNR) as the main evaluation metrics, which are also calculated on $256\times 256$ images. More details can be found in Supplementary.

\vspace{-0.2cm}
\subsection{Comparison with Existing Methods}
Both traditional methods~\cite{lalonde2007using,xue2012understanding} and deep learning based methods~\cite{zhu2015learning,tsai2017deep,xiaodong2019improving,DoveNet2020} are included for quantitative comparisons. Following \cite{tsai2017deep,DoveNet2020}, we train the model on the merged training sets of four sub-datasets in iHarmony4. The trained model is evaluated on each test set and the merged test set as well. Table \ref{tab:baselines} shows the quantitative results of different harmonization methods. The S$^2$AM~\cite{xiaodong2019improving} model is realized with recently released code and the other results of previous baselines are directly copied from \cite{DoveNet2020}. From Table \ref{tab:baselines}, we can observe that our method not only significantly exceeds traditional methods on all sub-datasets, but also outperforms deep learning based approaches on the whole test set. 
Besides, following~\cite{DoveNet2020}, we also investigate the the MSE and foreground MSE (fMSE) on the test images in different foreground ratio ranges (\emph{e.g.}, $5\%\sim 15\%$) in the Supplementary.
%
\vspace{-0.3cm}
\subsection{Ablation Studies}\label{sec:ablate}

We analyze the impact of hyper-parameters (\emph{i.e.}, the margin $m$ in Eqn. (\ref{eqn:triplet_fb})(\ref{eqn:triplet_ff}), $\lambda$ in Eqn. (\ref{eqn:total_loss}) and the domain code dimension $L$) in our method. We also investigate the impact of each type of network input and ablate each type of triplet loss to prove the necessity of mask, background domain code, and two triplet losses. Due to space limitation, we leave the detailed experimental results to Supplementary.

\vspace{-0.3cm}
\subsection{Domain Code Analyses}\label{sec:domain_code}
Recall that we employ two triplet losses Eqn. (\ref{eqn:triplet_fb})(\ref{eqn:triplet_ff}) to regulate the domain code. To verify that the expected requirements are satisfied on the training set and generalizable to the test set, we conduct domain code analyses on both training set and test set. Since DoveNet employs a domain verification discriminator to extract foreground and background domain representations, DoveNet is also included for comparison. As defined in Section~\ref{sec:bargainnet},  an image triplet contains a composite image, its ground-truth real image, and its harmonized output.
We calculate the ratio of training/testing image triplets which satisfy $d(\hat z_f, z_b)<d(\hat z_f, \tilde z_f)$ (\emph{resp.}, $d(z_f, \hat z_f)<d(z_f, \tilde z_f)$)  corresponding to  Eqn. (\ref{eqn:triplet_fb}) (\emph{resp.}, Eqn. (\ref{eqn:triplet_ff})) for both DoveNet and our method. For brevity, we use $d_{x,y}$ to denote $d(z_x, z_y)$, as shown in Table~\ref{tab:domain_code}. 

More generally, in an image triplet, the background, the real foreground, and the harmonized foreground belong to the same domain, while the composite foreground belongs to another domain. Considering that the distance between cross-domain regions should be larger than the distance between same-domain regions, we could construct $6$ groups of (anchor, positive, negative) in the form of triplet loss, leading to $6$ requirements: $d_{b,f}<d_{b,\tilde f}$, $d_{b,\hat f}<d_{b,\tilde f}$, $d_{f,\hat f}<d_{f,\tilde f}$, $d_{\hat f, f}<d_{\hat f,\tilde f}$, $d_{\hat f, b}<d_{\hat f,\tilde f}$, and $d_{f, b}<d_{f,\tilde f}$. 
The verification results of each individual requirement and all requirements for DoveNet and our method are summarized in Table~\ref{tab:domain_code}. We can observe the high ratio of training/testing image triplets that satisfy each individual requirement for our method. Moreover, most training/testing image triplets satisfy all six requirements at the same time, which implies that compared with DoveNet, our domain code extractor can indeed extract the domain code which contains domain information as expected.

\vspace{-0.5cm}
\begin{figure*}[ht]
\setlength{\abovecaptionskip}{0.1cm}
\setlength{\belowcaptionskip}{-0.4cm}
\centering
  \includegraphics[width=0.90\linewidth]{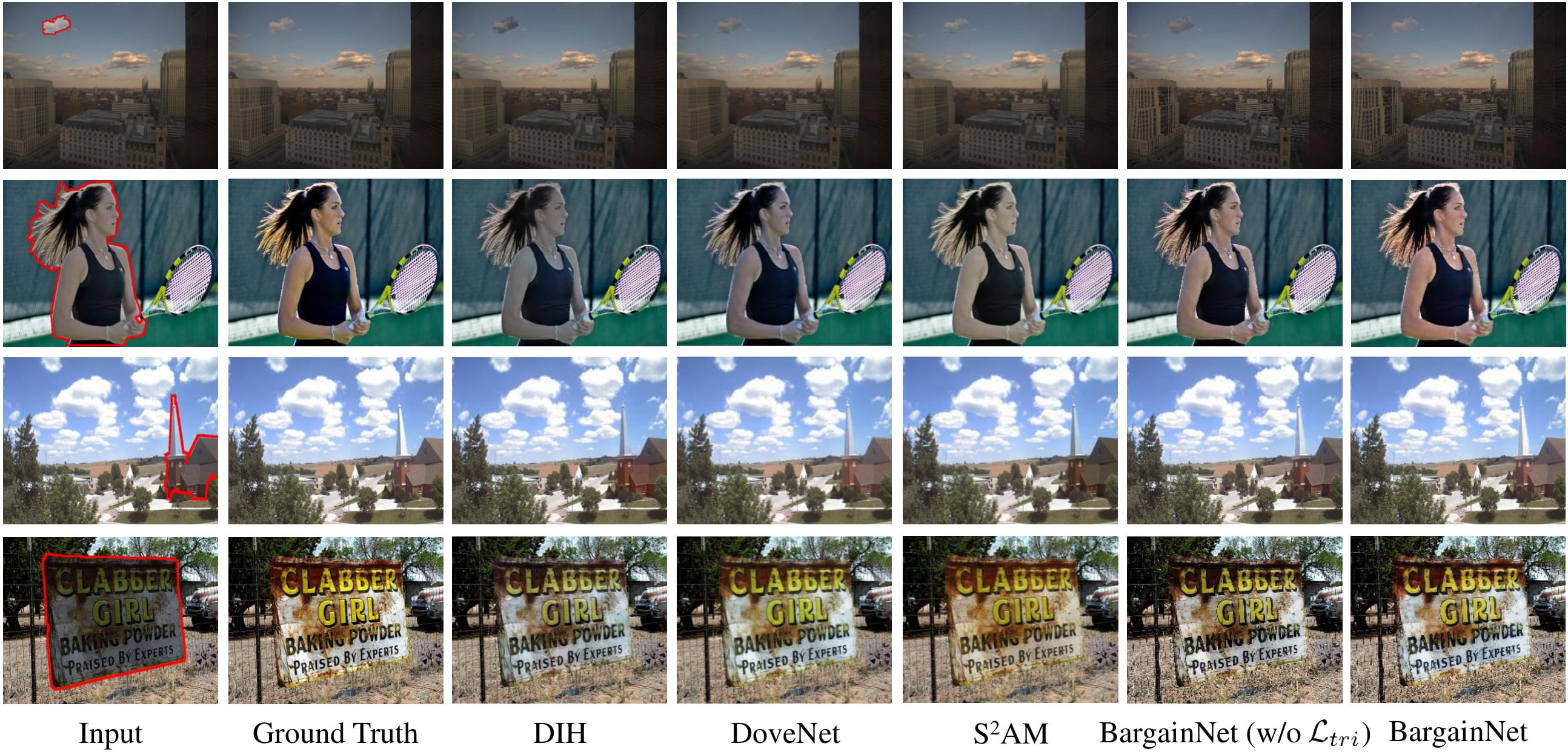}
  \caption[]{Example results of baselines and our method on four sub-datasets. From top to bottom, we show one example from HAdobe5k, HCOCO, Hday2night, and HFlickr sub-dataset respectively. From left to right, we show the input composite image, the ground-truth real image, and the results of  DIH~\cite{tsai2017deep}, DoveNet~\cite{DoveNet2020}, S$^2$AM~\cite{xiaodong2019improving}, our special case BargainNet (w/o $\mathcal{L}_{tri}$) and our proposed BargainNet respectively. The foregrounds are highlighted with red border lines for clarity.}
  \label{fig:samples}
\end{figure*}

\vspace{0.1cm}
\subsection{Qualitative Analyses}\label{sec:qualitative}
Given an input composite image from the test set, the harmonized outputs generated by DIH~\cite{tsai2017deep}, DoveNet~\cite{DoveNet2020}, S$^2$AM~\cite{xiaodong2019improving}, BargainNet (w/o $\mathcal{L}_{tri}$) and BargainNet are shown in Fig. \ref{fig:samples}. BargainNet (w/o $\mathcal{L}_{tri}$) is a special case without triplet losses. Compared with other baselines, BargainNet could generate more favorable results with consistent foreground and background, which are visually closer to the ground-truth real images. Besides, by comparing BargainNet with BargainNet (w/o $\mathcal{L}_{tri}$), we can observe that the generated outputs of BargainNet are more harmonious after using triplet losses, which provides an intuitive demonstration that triplet losses contribute to more effective domain code extraction.

In the real-world applications, given a real composite image, there is no ground-truth as the synthesized composite, so it is infeasible to evaluate the model performance quantitatively using MSE or PSNR. Following \cite{tsai2017deep,xiaodong2019improving,DoveNet2020}, we conduct user study on 99 real composite images~\cite{tsai2017deep}, in which we compare our BargainNet with all the other deep learning based methods. The details of user study and harmonization results can be found in the Supplementary.

\vspace{-0.2cm}
\subsection{Background Harmonization and Inharmony Level Prediction}\label{extension}

By inverting the mask fed into the generator and the domain code extractor in the testing stage, our BargainNet could be easily applied to background harmonization, which means adjusting the background to make it compatible with the foreground. We show our background harmonization results and compare with other deep learning based methods in Supplementary. 

Besides, one byproduct of our method is predicting the inharmony level of a composite image, which reflects how inharmonious this composite image is.
In particular, based on the extracted domain codes of the foreground region and background region, we can assess the inharmony level by calculating the Euclidean distance between two domain codes. The detailed inharmony level analyses are also left to Supplementary due to space limitation.

\vspace{-0.2cm}
\section{Conclusion}
In this work, we have proposed to formulate image harmonization as background-guided domain translation, which provides a new perspective for image harmonization. We have also presented BargainNet, a novel network that leverages the background domain code for foreground harmonization. Experimental results have shown that our method performs favorably on both the synthesized dataset iHarmony4 and real composite images.

\vspace{-0.3cm}
\section{Acknowledgement}
The work is supported by the National Key R\&D Program of China (2018AAA0100704) and is partially sponsored by National Natural Science Foundation of China (Grant No.61902247) and Shanghai Sailing Program (19YF1424400). 

\begin{small}
\vspace{-0.2cm}
\bibliographystyle{IEEEbib}
\bibliography{egbib}

\begin{thebibliography}{10}

\bibitem{tsai2017deep}
Yi{-}Hsuan Tsai, Xiaohui Shen, Zhe Lin, Kalyan Sunkavalli, Xin Lu, and
  Ming{-}Hsuan Yang,
\newblock ``Deep image harmonization,''
\newblock in {\em CVPR}, 2017.

\bibitem{xiaodong2019improving}
Xiaodong Cun and Chi{-}Man Pun,
\newblock ``Improving the harmony of the composite image by spatial-separated
  attention module,''
\newblock {\em {IEEE} Trans. Image Process.}, 2020.

\bibitem{DoveNet2020}
Wenyan Cong, Jianfu Zhang, Li~Niu, Liu Liu, Zhixin Ling, Weiyuan Li, and Liqing
  Zhang,
\newblock ``{DoveNet}: Deep image harmonization via domain verification,''
\newblock in {\em CVPR}, 2020.

\bibitem{lalonde2007using}
Jean{-}Fran{\c{c}}ois Lalonde and Alexei~A. Efros,
\newblock ``Using color compatibility for assessing image realism,''
\newblock in {\em ICCV}, 2007.

\bibitem{xue2012understanding}
Su~Xue, Aseem Agarwala, Julie Dorsey, and Holly~E. Rushmeier,
\newblock ``Understanding and improving the realism of image composites,''
\newblock {\em {ACM} Transactions on Graphics}, 2012.

\bibitem{multi-scale}
Kalyan Sunkavalli, Micah~K. Johnson, Wojciech Matusik, and Hanspeter Pfister,
\newblock ``Multi-scale image harmonization,''
\newblock {\em {ACM} Transactions on Graphics}, 2010.

\bibitem{pixelGAN}
Phillip Isola, Jun{-}Yan Zhu, Tinghui Zhou, and Alexei~A. Efros,
\newblock ``Image-to-image translation with conditional adversarial networks,''
\newblock in {\em CVPR}, 2017.

\bibitem{cycleGAN}
Jun{-}Yan Zhu, Taesung Park, Phillip Isola, and Alexei~A. Efros,
\newblock ``Unpaired image-to-image translation using cycle-consistent
  adversarial networks,''
\newblock in {\em ICCV}, 2017.

\bibitem{starGan}
Yunjey Choi, Min{-}Je Choi, Munyoung Kim, Jung{-}Woo Ha, Sunghun Kim, and
  Jaegul Choo,
\newblock ``{StarGAN}: Unified generative adversarial networks for multi-domain
  image-to-image translation,''
\newblock in {\em CVPR}, 2018.

\bibitem{lee2020drit++}
Hsin-Ying Lee, Hung-Yu Tseng, Qi~Mao, Jia-Bin Huang, Yu-Ding Lu, Maneesh Singh,
  and Ming-Hsuan Yang,
\newblock ``Drit++: Diverse image-to-image translation via disentangled
  representations,''
\newblock {\em International Journal of Computer Vision}, 2020.

\bibitem{choi2020stargan}
Yunjey Choi, Youngjung Uh, Jaejun Yoo, and Jung{-}Woo Ha,
\newblock ``{StarGAN v2}: Diverse image synthesis for multiple domains,''
\newblock in {\em CVPR}, 2020.

\bibitem{f2gan}
Yan Hong, Li~Niu, Jianfu Zhang, Weijie Zhao, Chen Fu, and Liqing Zhang,
\newblock ``{F2GAN}: Fusing-and-filling gan for few-shot image generation,''
\newblock in {\em MM}, 2020.

\bibitem{matchinggan}
Yan Hong, Li~Niu, Jianfu Zhang, and Liqing Zhang,
\newblock ``{Matchinggan}: Matching-based few-shot image generation,''
\newblock in {\em ICME}, 2020.

\bibitem{anokhin2020high}
Ivan Anokhin, Pavel Solovev, Denis Korzhenkov, Alexey Kharlamov, Taras
  Khakhulin, Aleksei Silvestrov, Sergey Nikolenko, Victor Lempitsky, and Gleb
  Sterkin,
\newblock ``High-resolution daytime translation without domain labels,''
\newblock in {\em CVPR}, 2020.

\bibitem{wang2019example}
Miao Wang, Guo-Ye Yang, Ruilong Li, Run-Ze Liang, Song-Hai Zhang, Peter~M Hall,
  and Shi-Min Hu,
\newblock ``Example-guided style-consistent image synthesis from semantic
  labeling,''
\newblock in {\em CVPR}, 2019.

\bibitem{Liu2018}
Guilin Liu, Fitsum~A. Reda, Kevin~J. Shih, Ting{-}Chun Wang, Andrew Tao, and
  Bryan Catanzaro,
\newblock ``Image inpainting for irregular holes using partial convolutions,''
\newblock in {\em ECCV}, 2018.

\bibitem{reinhard2001color}
Erik Reinhard, Michael Ashikhmin, Bruce Gooch, and Peter Shirley,
\newblock ``Color transfer between images,''
\newblock {\em {IEEE} Computer Graphics and Applications}, 2001.

\bibitem{colorharmonization}
Daniel Cohen{-}Or, Olga Sorkine, Ran Gal, Tommer Leyvand, and Ying{-}Qing Xu,
\newblock ``Color harmonization,''
\newblock {\em {ACM} Transactions on Graphics}, 2006.

\bibitem{poisson}
Patrick P{\'{e}}rez, Michel Gangnet, and Andrew Blake,
\newblock ``Poisson image editing,''
\newblock {\em {ACM} Transactions on Graphics}, 2003.

\bibitem{zhu2015learning}
Jun{-}Yan Zhu, Philipp Kr{\"{a}}henb{\"{u}}hl, Eli Shechtman, and Alexei~A.
  Efros,
\newblock ``Learning a discriminative model for the perception of realism in
  composite images,''
\newblock in {\em ICCV}, 2015.

\bibitem{luan2018deep}
Fujun Luan, Sylvain Paris, Eli Shechtman, and Kavita Bala,
\newblock ``Deep painterly harmonization,''
\newblock {\em Computer Graphics Forum}, 2018.

\bibitem{shaham2019singan}
Tamar~Rott Shaham, Tali Dekel, and Tomer Michaeli,
\newblock ``{SinGAN}: Learning a generative model from a single natural
  image,''
\newblock in {\em ICCV}, 2019.

\bibitem{huang2018munit}
Xun Huang, Ming{-}Yu Liu, Serge~J. Belongie, and Jan Kautz,
\newblock ``Multimodal unsupervised image-to-image translation,''
\newblock in {\em ECCV}, 2018.

\bibitem{ma2018exemplar}
Liqian Ma, Xu~Jia, Stamatios Georgoulis, Tinne Tuytelaars, and Luc~Van Gool,
\newblock ``Exemplar guided unsupervised image-to-image translation with
  semantic consistency,''
\newblock in {\em ICLR}, 2019.

\bibitem{zhu2017toward}
Jun-Yan Zhu, Richard Zhang, Deepak Pathak, Trevor Darrell, Alexei~A Efros,
  Oliver Wang, and Eli Shechtman,
\newblock ``Toward multimodal image-to-image translation,''
\newblock in {\em NeurIPS}, 2017.

\end{thebibliography}


\begin{thebibliography}{10}

\bibitem{DoveNet2020}
Wenyan Cong, Jianfu Zhang, Li~Niu, Liu Liu, Zhixin Ling, Weiyuan Li, and Liqing
  Zhang,
\newblock ``{DoveNet}: Deep image harmonization via domain verification,''
\newblock in {\em CVPR}, 2020.

\bibitem{lin2014microsoft}
Tsung{-}Yi Lin, Michael Maire, Serge~J. Belongie, James Hays, Pietro Perona,
  Deva Ramanan, Piotr Doll{\'{a}}r, and C.~Lawrence Zitnick,
\newblock ``Microsoft {COCO:} common objects in context,''
\newblock in {\em ECCV}, 2014.

\bibitem{bychkovsky2011learning}
Vladimir Bychkovsky, Sylvain Paris, Eric Chan, and Fr{\'{e}}do Durand,
\newblock ``Learning photographic global tonal adjustment with a database of
  input / output image pairs,''
\newblock in {\em CVPR}, 2011.

\bibitem{ade20k}
Bolei Zhou, Hang Zhao, Xavier Puig, Tete Xiao, Sanja Fidler, Adela Barriuso,
  and Antonio Torralba,
\newblock ``Semantic understanding of scenes through the {ADE20K} dataset,''
\newblock {\em Int. J. Comput. Vis.}, 2019.

\bibitem{zhou2016evaluating}
Hao Zhou, Torsten Sattler, and David~W. Jacobs,
\newblock ``Evaluating local features for day-night matching,''
\newblock in {\em ECCV}, 2016.

\bibitem{lalonde2007using}
Jean{-}Fran{\c{c}}ois Lalonde and Alexei~A. Efros,
\newblock ``Using color compatibility for assessing image realism,''
\newblock in {\em ICCV}, 2007.

\bibitem{xue2012understanding}
Su~Xue, Aseem Agarwala, Julie Dorsey, and Holly~E. Rushmeier,
\newblock ``Understanding and improving the realism of image composites,''
\newblock {\em {ACM} Transactions on Graphics}, 2012.

\bibitem{zhu2015learning}
Jun{-}Yan Zhu, Philipp Kr{\"{a}}henb{\"{u}}hl, Eli Shechtman, and Alexei~A.
  Efros,
\newblock ``Learning a discriminative model for the perception of realism in
  composite images,''
\newblock in {\em ICCV}, 2015.

\bibitem{tsai2017deep}
Yi{-}Hsuan Tsai, Xiaohui Shen, Zhe Lin, Kalyan Sunkavalli, Xin Lu, and
  Ming{-}Hsuan Yang,
\newblock ``Deep image harmonization,''
\newblock in {\em CVPR}, 2017.

\bibitem{xiaodong2019improving}
Xiaodong Cun and Chi{-}Man Pun,
\newblock ``Improving the harmony of the composite image by spatial-separated
  attention module,''
\newblock {\em {IEEE} Trans. Image Process.}, 2020.

\bibitem{bradley1952rank}
Ralph~Allan Bradley and Milton~E Terry,
\newblock ``Rank analysis of incomplete block designs: I. the method of paired
  comparisons,''
\newblock {\em Biometrika}, 1952.

\bibitem{lai2016comparative}
Wei{-}Sheng Lai, Jia{-}Bin Huang, Zhe Hu, Narendra Ahuja, and Ming{-}Hsuan
  Yang,
\newblock ``A comparative study for single image blind deblurring,''
\newblock in {\em CVPR}, 2016.

\end{thebibliography}
\end{small}

\end{document}


\sloppy

\def\x{{\mathbf x}}
\def\L{{\cal L}}

\title{Supplementary Material for BargainNet: Background-Guided Domain Translation for Image Harmonization}
%
\name{Wenyan Cong, Li Niu\textsuperscript{*}\thanks{\textsuperscript{*}Corresponding author.}, Jianfu Zhang, Jing Liang, Liqing Zhang}
\address {MoE Key Lab of Artificial Intelligence, Department of Computer Science and Engineering \\
Shanghai Jiao Tong University, Shanghai, China \\ 
\{plcwyam17320, ustcnewly, c.sis, leungjing\}@sjtu.edu.cn, zhang-lq@cs.sjtu.edu.cn.}

\maketitle

In this Supplementary file, we will introduce the details of iHarmony4 dataset and our network implementation in Section~\ref{sec:dataset}, \ref{sec:implementation}. Then, we will show the comparison with existing baselines in different foreground ratio ranges in Section~\ref{sec:fmse}. And we will analyze the impact of different hyper-parameters and ablate each part of the input and each type of triplet loss in Section~\ref{sec:ablation_detail}. Besides, we will introduce more details of user study conducted on real composite images and show some harmonized results of deep learning based methods on real composite images in Section~\ref{sec:user_study}. Finally, we will exhibit the background harmonization results of deep learning based methods in Section~\ref{sec:bg_harmonization}, and analyze the inharmony level prediction of our method in Section~\ref{sec:inharmony_level}.

\section{Dataset Statistics}\label{sec:dataset}
The iHarmony4 dataset contributed by \cite{DoveNet2020} is composed of pairs of synthesized composite images and the ground-truth real images. iHarmony4 consists of 4 sub-datasets: HCOCO, HAdobe5k, HFlickr, and Hday2night.

\textbf{HCOCO} sub-dataset is synthesized based on the merged training and test splits of Microsoft COCO \cite{lin2014microsoft}, containing 38545 training and 4283 test pairs of composite and real images. In HCOCO, the composite images are synthesized from real images and the foreground of composite image is adjusted by transferring the color from another foreground object of the same class in COCO using color mapping functions.

\textbf{HAdobe5k} sub-dataset is generated based on MIT-Adobe FiveK dataset \cite{bychkovsky2011learning}, containing 19437 training and 2160 test pairs of composite and real images. The composite image is generated by exchanging the manually segmented foreground between the real image and its five different renditions.

\textbf{HFlickr} sub-dataset is synthesized based on the crawled images from Flickr, containing 7449 training and 828 test pairs of composite and
real images. The composite images are synthesized similarly to HCOCO, except that the reference foreground is selected from ADE20K~\cite{ade20k} using the dominant category labels generated by pre-trained scene parsing model.

\textbf{Hday2night} sub-dataset is generated based on day2night \cite{zhou2016evaluating}, containing 311 training and 133 test pairs of composite and real images. The composite images are synthesized similarly to HAdobe5k, where the foreground is exchanged between images captured in different conditions.

\section{Implementation Details}\label{sec:implementation}

Our network is trained on ubuntu 16.04 LTS operation system, with 64GB memory, Intel Core i7-8700K CPU, and two GeForce GTX 1080 Ti GPUs. The network is implemented using Pytorch 1.4.0 and the weight is initialized with values drawn from the normal distribution $\mathcal{N}({mean}=0.0, {std}^2=0.02)$.

The domain code extractor $E$ is comprised of five partial convolutional layers with kernel size 3 and stride 2, an adaptive average pooling layer, and a convolutional layer with kernel size 1 and stride 1. Each of the partial convolutional layers is followed by ReLU and batch normalization except the last one.

\section{Comparison with Existing Methods}\label{sec:fmse}

Following DoveNet~\cite{DoveNet2020}, we also report the MSE and foreground MSE (fMSE) on the test images in different foreground ratio ranges (\emph{e.g.}, $5\%\sim 15\%$). The foreground ratio means the area of the foreground over the area of the whole image. Foreground MSE (fMSE) is MSE calculated only in the foreground region.
As shown in Table~\ref{tab:ablate_ratio}, our method outperforms all the baselines in each foreground ratio range, which demonstrates the robustness of our proposed method.

\section{Ablation Studies}\label{sec:ablation_detail}

\subsection{Hyper-parameter Analyses}\label{sec:parameter}
\begin{figure}[tp!]
\begin{center}
\includegraphics[width=1\linewidth]{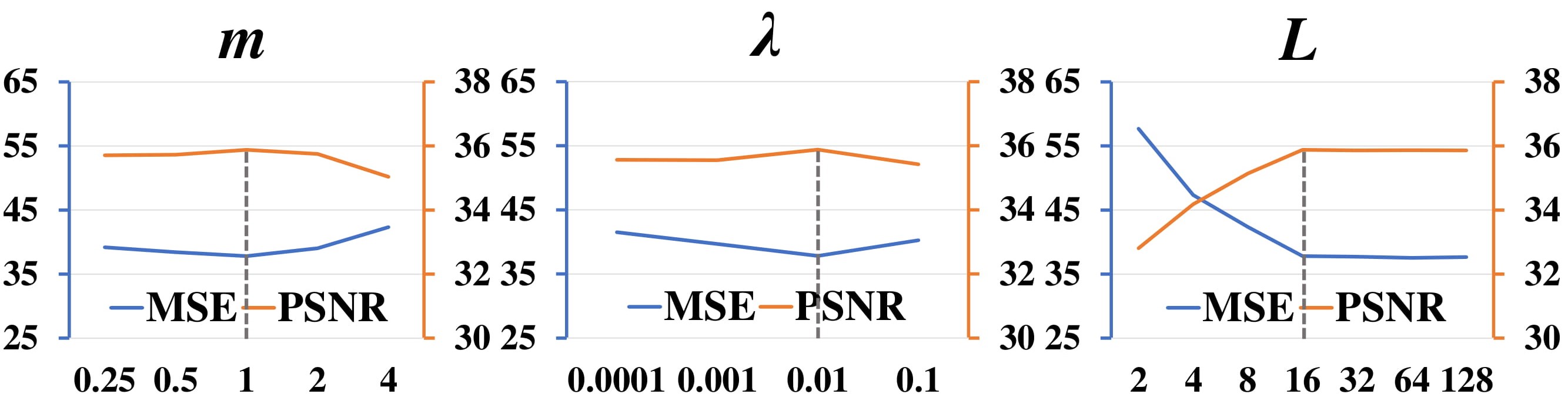}
\end{center}
   \caption{Impact of hyper-parameters, including the margin $m$ in Eqn. (1)(2), $\lambda$ in Eqn. (3), and the domain code dimension $L$. The gray dotted line indicates the default value of each hyper-parameter.}
\label{fig:hyperparameter}
\end{figure}

We investigate the impact of three hyper-parameters: the margin $m$ in Eqn. (1)(2), $\lambda$ in Eqn. (3), and the domain code dimension $L$. In Fig.~\ref{fig:hyperparameter}, we plot the performance by varying each hyper-parameter while keeping the other hyper-parameters fixed. It can be seen that our method is robust with $m$ (\emph{resp.}, $\lambda$) in a reasonable range $[2^{-2},2^{2}]$ (\emph{resp.}, $[10^{-4},10^{-1}]$). With the domain code dimension increasing to 16, the performance improves obviously. When $L$ is larger than 16, the performance increases marginally, but more training resources are in demand. So 16-dimensional domain code is a cost-effective choice. 

\subsection{Input Design Choices}\label{sec:ablate}

As described in Section 3.2 in the main text, we concatenate the composite image, foreground mask, and background domain code map as input for our generator $G$. Now we investigate the impact of each type of input and report the results in Table~\ref{tab:ablate_loss}.
When we only use composite image and foreground mask as input (row 1), it is exactly the same as the attention-enhanced U-net introduced in \cite{DoveNet2020}. After adding the background domain code to the input (row 2), the performance is significantly boosted, which demonstrates that background domain code can provide useful guidance for foreground harmonization. We further apply our proposed two triplet losses to regulate the domain code (row 3), which brings in extra performance gain. This is because that the triplet losses impose reasonable constraints for better domain code extraction. 

Besides, when we replace the background domain code with the domain code extracted from the whole composite image (row 8), the harmonization performance is degraded by a large margin (row 8 \emph{v.s.} row 3). This is because in composite image, the foreground and background are captured in different conditions and belong to two different domains. Domain code extracted from the whole image will mislead the harmonization with undesirable foreground domain information.

In addition, we also investigate the case in which we only feed the generator with composite image and the background domain code map while removing the foreground mask from input. No matter using triplet losses (row 6) or not (row 7), the performance is significantly degraded after removing the foreground mask (row 6 \emph{v.s.} row 3, row 7 \emph{v.s.} row 2), probably because the foreground mask emphasizes foreground translation and enables the foreground to borrow information from background.

\subsection{Loss Design Choices}
Besides, we also ablate each type of triplet loss (row 4 and row 5) in Table~\ref{tab:ablate_loss}. The results demonstrate that each type of triplet loss is helpful, and two types of triplet losses can collaborate with each other to achieve further improvement.

\begin{table*}[tb]
\setlength{\abovecaptionskip}{0.1cm}
\centering
\begin{tabular}{|l|c|c|c|c|c|c|c|c|}
\hline
\multicolumn{1}{|c|}{Foreground ratios} & \multicolumn{2}{c|}{$0\%\sim 5\%$} & \multicolumn{2}{c|}{$5\%\sim 15\%$} & \multicolumn{2}{c|}{$15\%\sim 100\%$}  & \multicolumn{2}{c|}{$0\%\sim 100\%$}\\ \hline
\multicolumn{1}{|c|}{Evaluation metric}  & MSE$\downarrow$  & fMSE$\downarrow$   & MSE$\downarrow$  & fMSE$\downarrow$ & MSE$\downarrow$  & fMSE$\downarrow$ & MSE$\downarrow$  & fMSE$\downarrow$   \\ \hline
\multicolumn{1}{|c|}{Input composite}  & 28.51  & 1208.86   & 119.19  & 1323.23 & 577.58  & 1887.05 & 172.47 & 1387.30   \\ \hline
\multicolumn{1}{|c|}{Lalonde and Efros\cite{lalonde2007using}}  & 41.52  & 1481.59   & 120.62  & 1309.79 & 444.65  & 1467.98 & 150.53 & 1433.21  \\ \hline
\multicolumn{1}{|c|}{Xue \emph{et al.}\cite{xue2012understanding}}  & 31.24  & 1325.96   & 132.12  & 1459.28 & 479.53  & 1555.69 & 155.87 & 1411.40   \\ \hline
\multicolumn{1}{|c|}{Zhu \emph{et al.}\cite{zhu2015learning}}  & 33.30  & 1297.65   & 145.14  & 1577.70 & 682.69  & 2251.76 & 204.77 & 1580.17  \\ \hline
\multicolumn{1}{|c|}{DIH~\cite{tsai2017deep}}  & 18.92  & 799.17   & 64.23  & 725.86 & 228.86  & 768.89 & 76.77 & 773.18  \\ \hline
\multicolumn{1}{|c|}{DoveNet~\cite{DoveNet2020}}  & 14.03  & 591.88 & 44.90  & 504.42 & 152.07  & 505.82 & 52.36 & 549.96  \\ \hline
\multicolumn{1}{|c|}{S$^2$AM~\cite{xiaodong2019improving}}  & 13.51  & 509.41 & 41.79 & 454.21 & 137.12  & 449.81 & 48.00 & 481.79  \\ \hline
\multicolumn{1}{|c|}{Ours}  & \bf10.55  & \bf450.33  & \bf32.13  & \bf359.49 & \bf109.23 & \bf353.84 & \bf37.82 & \bf 405.23 \\ \hline
\end{tabular}
\caption{MSE and foreground MSE (fMSE) of different methods in each foreground ratio range based on the whole test set. The best results are denoted in boldface.}
\label{tab:ablate_ratio}
\end{table*}

\begin{table}[tb]
\centering
\begin{tabular}{c|cc|cc|cc}
\toprule
\# & mask & $z_b$ & $\mathcal{L}_{f\hat f}$ & $\mathcal{L}_{\hat fb}$  & MSE $\downarrow$ & PSNR $\uparrow$ \\
\hline \hline 
1 & \checkmark & & & & 60.79 & 34.15 \\
2 &\checkmark& \checkmark& & & 43.70 & 35.43 \\
3 &\checkmark &\checkmark&\checkmark &\checkmark & 37.82 & 35.88 \\
4 &\checkmark&\checkmark&\checkmark &  & 41.03 & 35.47  \\
5 &\checkmark& \checkmark& &\checkmark & 41.71 & 35.50  \\
6 & & \checkmark &\checkmark &\checkmark& 115.48 & 31.94\\
7 & & \checkmark& & & 120.49 &31.89 \\
8 &\checkmark & $\circ$ &\checkmark &\checkmark &43.18 &35.50 \\ 
\bottomrule
\end{tabular}
\caption{Ablation studies on input format and triplet losses. ``mask" means foreground mask, $z_b$ denotes the background domain code, and $\circ$ means that we replace the background domain code with the domain code extracted from the whole composite image. Two triplet losses are $\mathcal{L}_{f\hat f}$ and $\mathcal{L}_{\hat fb}$. }
\label{tab:ablate_loss}
\end{table}

\begin{figure*}[t]
\centering
  \includegraphics[width=0.9\linewidth]{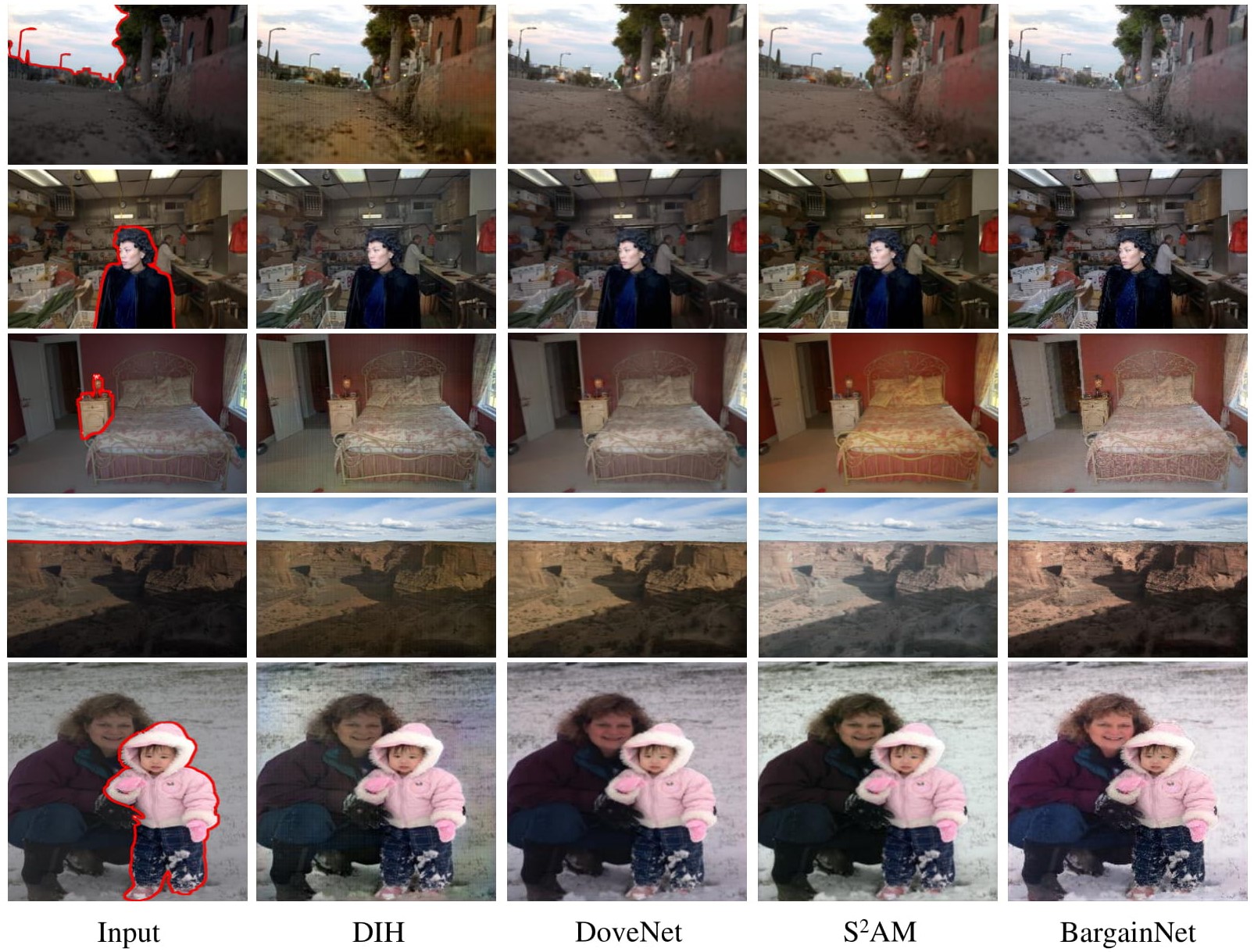}
  \caption[]{Example results of background harmonization. From left to right, we show the input composite image and the background harmonization results of DIH~\cite{tsai2017deep}, DoveNet~\cite{DoveNet2020}, S$^2$AM~\cite{xiaodong2019improving}, and our proposed BargainNet. For clarity, we highlight the unchanged foreground with red border lines.}
  \label{fig:bg_harm}
\end{figure*}

\begin{table}
\centering
\begin{tabular}{|c|c|}
\hline
Method & B-T score$\uparrow$\\\hline
Input composite & 0.357 \\ \hline
DIH~\cite{tsai2017deep}  & 0.813   \\ \hline
DoveNet \cite{DoveNet2020}  &  0.897 \\ \hline
S$^2$AM~\cite{xiaodong2019improving}  & 1.140 \\ \hline
Ours  & \bf 1.266  \\ \hline
\end{tabular}
\caption{B-T scores of deep learning based methods on $99$ real composite images provided in \cite{tsai2017deep}. }
\label{tab:BT_score}
\end{table}

\section{Results on Real Composite Images}\label{sec:user_study}

In reality, there is no ground-truth image for a given real composite image, whose foreground is cut from one image and pasted on another background image. In such a scenario, the foreground is not in the same location and its color distribution is vastly different from the background. Since it is infeasible to evaluate model performance quantitatively, following \cite{tsai2017deep,xiaodong2019improving,DoveNet2020}, we conduct user study on 99 real composite images released by \cite{tsai2017deep}. 
The perceptual evaluations in previous works~\cite{tsai2017deep,xiaodong2019improving,DoveNet2020} have shown that deep learning based methods are generally better than traditional methods, so we only compare our proposed BargainNet with deep learning based methods DIH~\cite{tsai2017deep}, DoveNet~\cite{DoveNet2020}, and S$^2$AM~\cite{xiaodong2019improving}. 

Similarly, given each composite image and its four harmonized outputs from four different methods, we can construct image pairs $(I_i, I_j)$ by randomly selecting from these five images $\{I_i|_{i=1}^5\}$. Hence, we can construct a large number of image pairs based on $99$ real composite images.
Each user involved in this subjective evaluation could see an image pair each time to decide which one looks more realistic. Considering the user bias, 22 users participate in the study in total, contributing 10835 pairwise results. With all pairwise results, we employ the Bradley-Terry (B-T) model  \cite{bradley1952rank,lai2016comparative} to obtain the global ranking of all methods and the results are reported in Table~\ref{tab:BT_score}. Our proposed BargainNet shows an advantage over other deep-based methods with the highest B-T score, which demonstrates that by explicitly using the background domain code as guidance, our method could generate more favorable results in real-world applications.

\begin{figure*}[tp!]
\begin{center}
\includegraphics[width=1\linewidth]{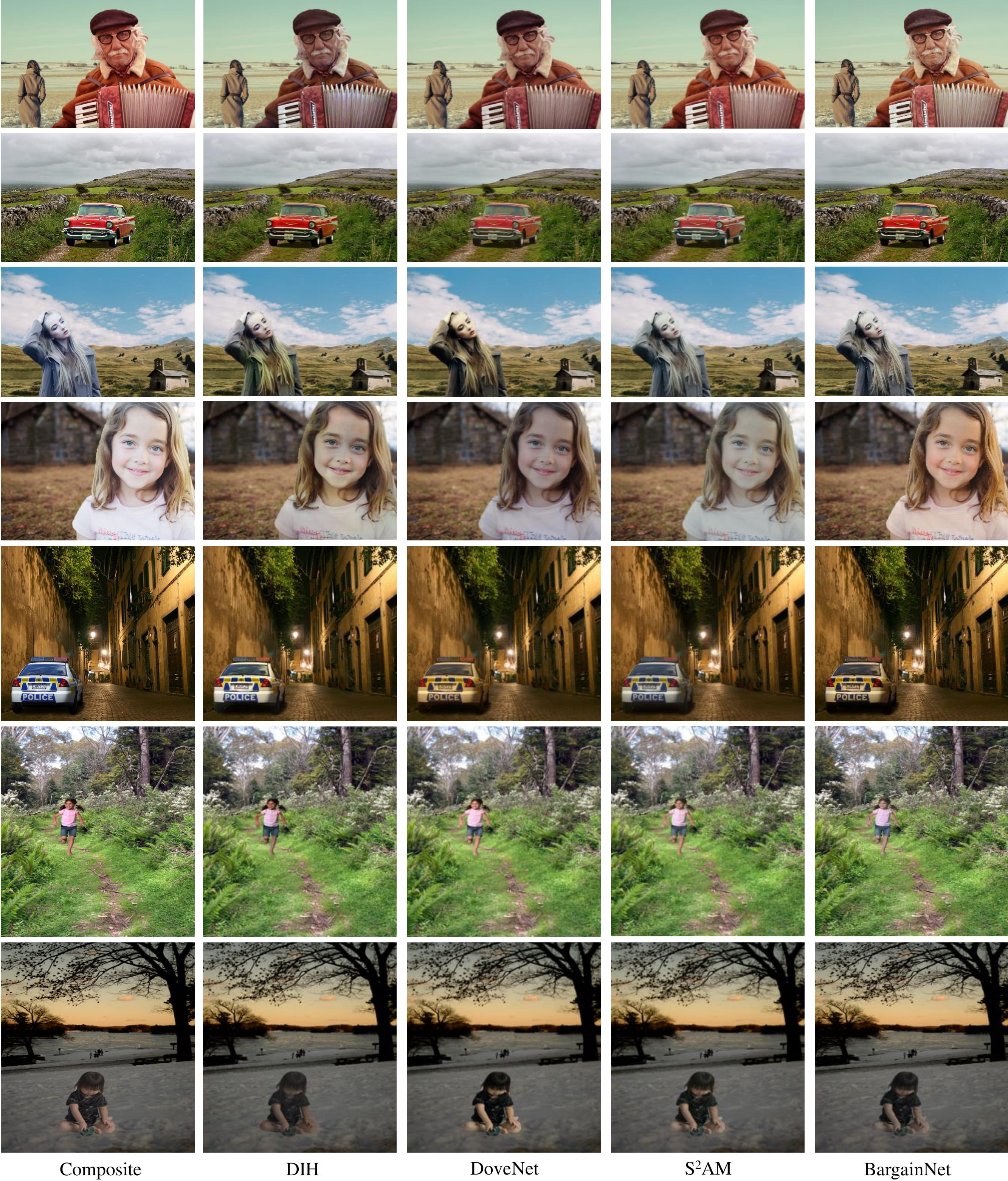}
\end{center}
   \caption{The harmonized results on real composite images, including three deep learning based methods and our proposed BargainNet.}
\label{fig:real_res}
\end{figure*}

In Fig.~\ref{fig:real_res}, we present some results of real composite images used in our user study. We compare the real composite images with harmonization results generated by our proposed method and other deep learning based methods, including DIH~\cite{tsai2017deep}, DoveNet~\cite{DoveNet2020}, and S$^2$AM~\cite{xiaodong2019improving}. Based on Fig.~\ref{fig:real_res}, we can see that our proposed method could generally produce satisfactory harmonized images compared to other deep learning based methods.

\begin{figure*}[tp!]
\begin{center}
\includegraphics[width=0.9\linewidth]{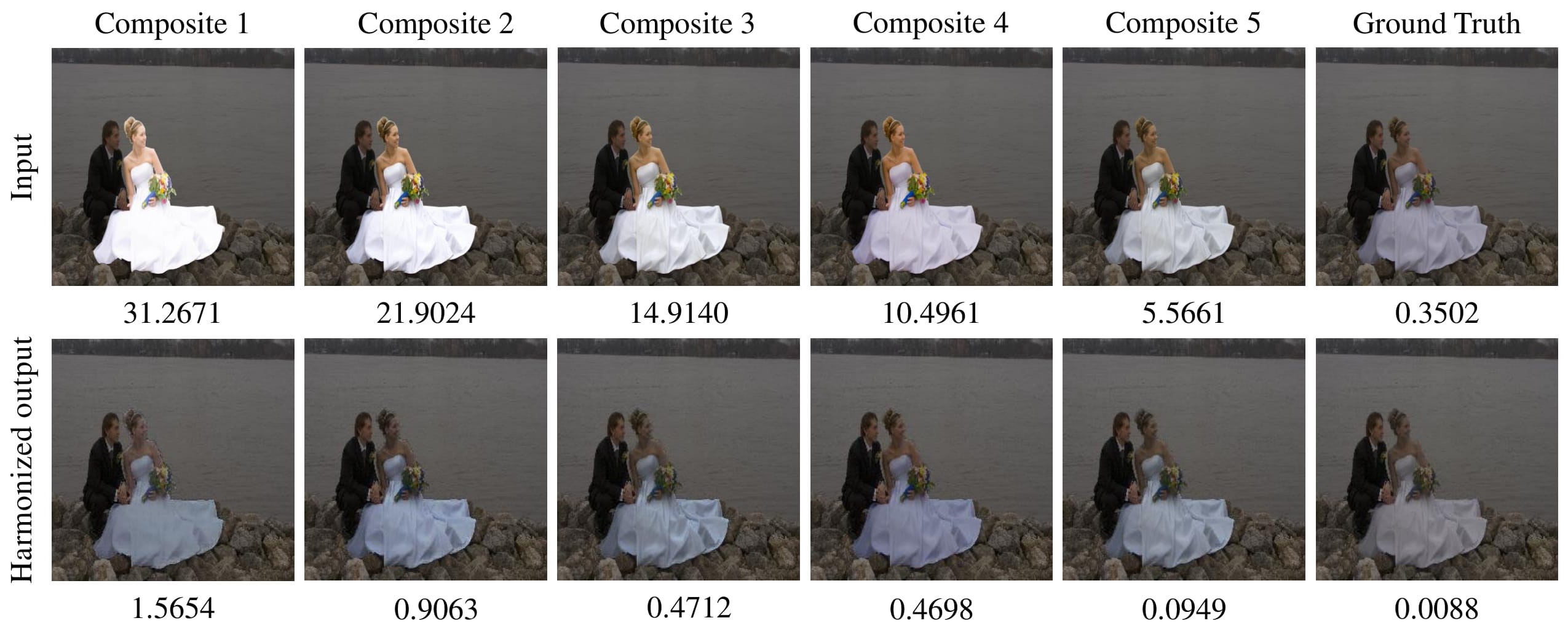}
\end{center}
   \caption{Examples of composite images with different inharmony levels. From top to bottom, we show the network input and the harmonized output of our BargainNet respectively. From left to right, we show the five composite images and the ground-truth real image. The number below each image is its  inharmony score.}
\label{fig:inharmonious}
\end{figure*}

\section{Generalization to Background Harmonization}\label{sec:bg_harmonization}

Interestingly, our method could also be used for background harmonization, in which the background is harmonized according to the foreground while the foreground remains unchanged. In particular, we can feed the composite image $\tilde I$, the background mask $\bar M$, and the composite foreground domain code $\tilde z_f$ into our generator $G$. In this way, the background region could be harmonized to the same domain as composite foreground, making the whole image harmonious as well. We show some background harmonization results of different deep learning based methods in Fig.~\ref{fig:bg_harm}. We can observe that our BargainNet is more capable of generating harmonious output. This observation is consistent with the observation in foreground harmonization, which demonstrates the remarkable generalizability and robustness of our method.

\section{Inharmony Level Prediction}\label{sec:inharmony_level}
Based on the extracted domain codes of foreground and background, we can predict the inharmony level of a composite image, reflecting to which extent the foreground is incompatible with the background. 

We conduct experiments on HAdobe5k sub-dataset, because each real image in MIT-Adobe FiveK dataset \cite{bychkovsky2011learning} has another five edited renditions of different styles. Given a real image, we can paste the foregrounds of five edited renditions on the background of real image, leading to five composite images with the same background yet different foregrounds in HAdobe5k. Therefore, when feeding the five composite images into $G$, the generated outputs are expected to be harmonized to the same ground-truth real image. Recall that in our BargainNet, we propose to use domain code extractor to extract the domain codes $\tilde z_f$ and $z_b$ for foreground and background respectively. So we can calculate the Euclidean distance $d(z_b, \tilde z_{f})$ as the inharmony score of a composite image, which reflects how inharmonious a composite image is. For the composite images with high inharmony scores, the foreground and the background are obviously inconsistent. After harmonization, the composite foreground is adjusted to be compatible with the background. Therefore, the inharmony score should become lower. 

In Fig.~\ref{fig:inharmonious}, we show one ground-truth real image with its five composite images from HAdobe5k sub-dataset, and report two inharmony scores of each image before and after harmonization. In the top row, we can observe that composite images whose foreground and background are obviously inconsistent have higher scores, while the ground-truth real image with consistent foreground and background has the lowest inharmony score. In the bottom row, after harmonization, as the foreground is translated to the same domain as background, the inharmony score of the harmonized output decreases dramatically. Interestingly, even for the ground-truth real image, harmonization using our method can further lower its inharmony score, probably because our network could make the foreground domain closer to the background.

Inharmony level provides an intuitive perspective for inharmony assessment, which is an enticing byproduct of our method and useful for harmonization related tasks. For example, given abundant composite images, we can first predict their inharmony levels and only harmonize those with high inharmony levels for computational efficiency.

\bibliographystyle{IEEEbib}
\bibliography{egbib}